\documentclass{article}
\usepackage[T1]{fontenc} 
\usepackage[utf8]{inputenc}
\usepackage[english]{babel}
\usepackage{xcolor}
\usepackage{graphicx}
\usepackage{caption}
\usepackage{subcaption}
\usepackage{geometry}
\usepackage{diagbox}
\usepackage{hyperref}
\usepackage{multicol}
\usepackage{float}
\restylefloat{table}
\usepackage{indentfirst}
\usepackage{bbm}
\DeclareUnicodeCharacter{00A0}{ }
\hypersetup{
  colorlinks = true, 
  urlcolor = blue, 
  linkcolor = blue, 
  citecolor = red 
}
\usepackage{csquotes}
\usepackage{amsmath}
\usepackage{amssymb}
\usepackage{microtype}
\usepackage{latexsym}
\usepackage[backend=biber, style=ieee]{biblatex}
\title{Multiple Samples Clustering}
\author{Xiang Wang, Tie Liu}
\date{July 2019}

\begin{document}
\maketitle

\begin{abstract}
    The clustering algorithms that view each object data as a single sample drawn from a certain distribution, Gaussian distribution, for example, has been a hot topic for decades. Many clustering algorithms: such as \textit{k}-means and spectral clustering, are proposed based on the single sample assumption. However, in real life, each input object can usually be the multiple samples drawn from a certain hidden distribution. The traditional clustering algorithms cannot handle such a situation. This calls for the multiple sample clustering algorithm. But the traditional multiple sample clustering algorithms can only handle scalar samples or samples from Gaussian distribution. This constrains the application field of multiple sample clustering algorithms. In this paper, we purpose a general framework for multiple sample clustering. Various algorithms can be generated by this framework. We apply two specific cases of this framework: Wasserstein distance version and Bhattacharyya distance version on both synthetic data and stock price data. The simulation results show that the sufficient statistic can greatly improve the clustering accuracy and  stability. 
\end{abstract}
\textbf{keywords: Gaussian distribution, Wasserstein distance, Bhattacharyya distance, multiple sample clustering, distribution information}

\section{Introduction}
Multiple sample data is one kind of specific data structure in clustering. Suppose there are different groups of vectors. The multiple sample clustering tries to cluster the groups rather than the vectors themselves. Most of the clustering algorithms e.g. \textit{k}-means, linear discriminant analysis, principal component analysis, etc, view each input object as a single sample drawn from univariate(or multivariate) Gaussian distribution.\par
Multiple samples clustering has a lot of applications in daily life. In video rating, for example, each video can be rated by different viewers and the rate scores of each video can be viewed as the multiple samples drawn from the same distribution. The clustering of different videos can help the video producers(such as Netflix) to recommend the most relevant movies(or ads) to the specified users\cite{davidson2010youtube,mei2009automatic,yang2007online}. The quality of products coming from the same batch can also be seen as the multiple samples drawn from a hidden distribution. If we can cluster the same batches together, the manager can use this information to improve the production process\cite{undey2003online,russell1998recursive}. What's more, the time-series data, such as sensor networks\cite{bandyopadhyay2003energy}, or the stock prices at different times\cite{harris1991stock}, can also be viewed as the multiple samples from a hidden distribution. The clustering of sensor networks can find the relevance between sensors and reduce the communication cost of different sensors. Clustering stocks can help us build the investment portfolio.\\

\textbf{Related work}
Multiple samples clustering is a subfield of unsupervised learning, which is a way to detect data partition without prior knowledge. There are two main ramifications of unsupervised learning, \textit{principal component}, and \textit{clustering}. \textit{Principal component}\cite{pearson1901liii} focuses on building linearly uncorrelated values based on a set of observations. This is not the main concern of our paper. Clustering aims at partitioning the whole observed set into separated subgroups. In the clustering field, scientists have been applying distribution information for decades. In 2006, Antonio Irpino\cite{irpino2006new} applied Wasserstein distance on histogram symbolic data clustering, for clustering different states in America based on temperatures of each month. In 2013, Claudia Canali and Riccardo Lancellotti\cite{canali2013automatic} applied Bhattacharyya distance into virtual machine clustering based on the behavior histogram of virtual machines, However, both algorithms are only useful for scalar samples. In 2007 Dillon\cite{davis2007differential} applied KL-divergence on multiple samples clustering under the Gaussian distribution assumption. The simulation result shows that distribution information can greatly improve the clustering accuracy than just using the mean vector information of each sample group, but KL divergence is not exactly a distance. Many graph-based clustering algorithms are inapplicable under this metric and this algorithm cannot fit all different multiple sample data structures. A general framework for multiple samples clustering remains undeveloped.\par
\textbf{Contribution and paper outline} In this paper, we propose a general framework for multiple sample clustering, various algorithms can be generated from this framework. An adapted algorithm called KL divergence++ is built based on Dhillion's work\cite{davis2007differential}, which can achieve higher clustering accuracy than the original one. Finally, we will compare the performance among five different algorithms to illustrate the importance of distribution information in multiple sample clustering.\par
This paper will be organized in the following structure: Section 2 includes the necessary notation definition and background knowledge.  Section 3 tries to build the model of the multiple samples clustering problem and propose the general framework to solve it. Two different clustering algorithms will be proposed in section 4 and the simulation results of both synthetic and real data will show in section 5. Future research directions and main results are contained in section 6.
 \cite{fortunato2016community}\cite{javed2018community}\cite{rossetti2018community}\par
\textbf{Necessary Notations} Although all symbols will be explicated when firstly referred to, we will talk about some basic notations in this paper for clarity.
\begin{table}[h]
    \caption{Essential Notations In This Paper}
    \centering
    \begin{tabular}{c l}
    \hline
    Mathematical Meaning & Symbols\\
    \hline
         vector &  bold lower case: $\mathbf{a,b,\dots}$\\
         matrix &  bold capital letter: $\mathbf{A,B,\dots}$\\
         spaces &  $\mathcal{X,Y\dots}$\\
         set & $\mathcal{A, C, S}$\\
         distribution & $s,\mu,\nu$\\
    \hline
    \end{tabular}
\end{table}
\section{Preliminary}
In this section, we are going to introduce some essential concepts and tools for multiple sample clustering. We all know the probability density function of univariate Gaussian distribution: $p(x|m,\sigma) = \frac{1}{\sqrt{2\pi\sigma^2}}e^{-\frac{(x-m)^2}{2\sigma^2}}$. Multivariate Gaussian distribution is a natural extension of it. Suppose a $d$-dimensional multivariate Gaussian distribution with mean vector $\mathbf{m}$ and covariance matrix $\mathbf{\Sigma}$. The probability density function is the following equation:
\begin{equation}
    p(\mathbf{x}|\mathbf{\mathbf{m},\mathbf{\Sigma}}) = \frac{1}{(2\pi)^{\frac{d}{2}|\mathbf{\Sigma}^|}}exp(-\frac{1}{2}(\mathbf{x}-\mathbf{m})^T\mathbf{\Sigma}^{-1}(\mathbf{x}-\mathbf{m}))
\end{equation} where $|\mathbf{\Sigma}|$ is the determinant of $\mathbf{\Sigma}$

\subsection{Distribution Metrics}
Mathematically, metric on a space $\mathcal{X}$ is a function:
$d: \mathcal{X}\times \mathcal{X} \rightarrow [0,+\infty)$ with the following four properties:
for elements $x,y,z \in \mathcal{X}$ and Cartesian product $\times$ between two spaces.
\begin{description}
    \item[non-negativity] $d(x,y)>0$
    \item[identity] $d(x,y) = 0 \Leftrightarrow x = y$
    \item[symmetry] $d(x,y) = d(y,x)$
    \item[subadditivity] $d(x,z)\leq d(x,y) + d(y,z)$
\end{description}
We will talk about two distribution metrics: Wasserstein distance and Bhattacharyya distance separately. KL divergence will show up in section 2.2.3 but it is not a distance metric since it does not obey the subadditivity rule. 
\subsubsection{Wasserstein Distance} 
Wasserstein distance can be retrospected to 1781 when Gaspard Monge made up the optimal transport object function to measure the effort to move one pile of dirt into another place with different shape\cite{monge1781memoire}. Wasserstein distance is also called the Earth movers' distance because of this history.
Let's set $\mathcal{X},\mathcal{Y}$ two spaces, $\mu$ and $\nu$ respectively two measurements defined on $\mathcal{X}$ and $\mathcal{Y}$ satisfying $\int_{\mathcal{X}}\mu(x)dx = 1,\int_{\mathcal{Y}}\nu(y)dy = 1$, and a map $T: \mathcal{X}\rightarrow \mathcal{Y}$ with $\mu(\mathbf{A}) = \nu(T(\mathbf{A}))\quad $for$ \quad \forall \mathbf{A}\subset\mathcal{X}$. The Wasserstein distance is defined as the minimum value of the cost function $c(x, y)$ via choosing the optimal map $T$
\begin{equation}
    D_W(\mu,\nu) = \min\limits_{T} \int_{\mathcal{X}} c(x, T(x))d\mu(x)dx \quad x\in\mathcal{X}, y\in\mathcal{Y}
\end{equation}
In practice, for vector $\mathbf{x}$ and $\mathbf{y}$ in Euclidean space, we always set $c(\mathbf{x},\mathbf{y}) = ||\mathbf{x}-\mathbf{y}||_2$ and the corresponded $D_{W,2}(\mu,\nu)$ is called 2-norm Wasserstein.\par
Although Wasserstein distance can measure the distance between continuous distributions, discrete distributions or between a continuous distribution and a discrete distribution, we only consider the Gaussian distribution for computation efficiency. Nevertheless, the framework proposed in this paper can be easily extended to different distributions.

Suppose two Gaussians $\mathbf{\mu}_1\sim\mathcal{N}(\mathbf{m}_1, \mathbf{\Sigma}_1)$ $\mathbf{\mu}_2\sim\mathcal{N}(\mathbf{m}_2,\mathbf{\Sigma}_2)$ where $\mathbf{m}_1$ and $\mathbf{m}_2$ are the mean vectors of two Gaussians and $\mathbf{\Sigma}_1$ and $\mathbf{\Sigma}_2$ are respectively covariance matrices. 
The Wasserstein distance of two $d$-dimension Gaussian distributions $\mathbf{\mu}_1, \mathbf{\mu}_2$ can be computed in the following equation:
\begin{equation}
    D_{W,2}(\mathbf{\mu}_1, \mathbf{\mu}_2) := ||\mathbf{m}_1 - \mathbf{m}_2||^2_2 + Tr(\mathbf{\Sigma}_1 + \mathbf{\Sigma}_2 - 2(\mathbf{\Sigma}_1^{1/2}\mathbf{\Sigma}_2\mathbf{\Sigma}_1^{1/2})) 
\end{equation}

\subsubsection{Bhattacharyya Distance}
Bhattacharyya Distance between two distributions $\mu$ and $\nu$ on the same domain $\mathcal{X}$ is 
\begin{equation}
    D_B(\mu,\nu) = \int_{x\in\mathcal{X}}\sqrt{\mu(x)\nu(x)}dx
\end{equation}
For the Bhattacharyya distance between Gaussian distributions $\mu\sim\mathcal{N}(\mathbf{m_1}, \mathbf{\Sigma_1})$, $\nu\sim\mathcal{N}(\mathbf{m_2},\mathbf{\Sigma_2})$
\begin{equation}
    D_B(\mu,\nu) = \frac{1}{8}(\mathbf{m}_1 - \mathbf{m}_2)^T\mathbf{\Sigma}^{-1}(\mathbf{m}_1 - \mathbf{m}_2) + \frac{1}{2}ln(\frac{|\mathbf{\Sigma}|}{\sqrt{|\mathbf{\Sigma}_1| |\mathbf{\Sigma}_2}|})
\end{equation}
where $\mathbf{\Sigma} = \frac{\mathbf{\Sigma}_1 + \mathbf{\Sigma}_2}{2}$
\subsubsection{KL divergence computation between Gaussians}
KL divergence between two distributions $\mu$ and $\nu$ is 
\begin{equation}
    D_{KL}(\mu||\nu) = \int_\mathcal{X}\mu(x)\log(\frac{\mu(x)}{\nu(x)})dx 
\end{equation} 
If $\mu\sim\mathcal{N}(\mathbf{m}_1,\mathbf{\Sigma}_1), \nu\sim\mathcal{N}(\mathbf{m}_2,\mathbf{\Sigma}_2)$ are two $d$-dimensional Gaussian distributions. The KL divergence between $\mu$ and $\nu$ is:
\begin{equation}
    D_{KL}(\mathbf{\mu}_1, \mathbf{\nu}_2) = \frac{1}{2}[\log\frac{|\mathbf{\Sigma}_2|}{|\mathbf{\Sigma}_1|} - d + Tr(\mathbf{\Sigma}_2^{-1}\mathbf{\Sigma}_1)+ (\mathbf{m}_2 - \mathbf{m}_1)^T\mathbf{\Sigma}_2^{-1}(\mathbf{m}_2-\mathbf{m}_1)]
\end{equation}
We notice that compared with Wasserstein distance and Bhattacharyya distance, KL divergence is not a real distance since KL divergence does not have the symmetry property. As a result, any graph partitioning algorithms, spectral clustering, for example, cannot be applied to the KL divergence adjacency matrix between Gaussians. 
\subsection{Spectral Clustering}
Spectral clustering is one of the graph partitioning algorithms. In graph partitioning problem, let $G = (\mathcal{V},\mathbf{W})$ be an undirected graph with vertex set $\mathcal{V} = \{v_1, v_2,\dots, v_n\}$ and similarity adjacency matrix $\mathbf{W}$. $w_{ij}\in\mathbf{W}$ represents the similarity between $v_i$ and $v_j$. Graph partitioning algorithms try to split $\mathcal{V}$ into non-overlapped parts $\{\mathcal{A}_1, \dots,  \mathcal{A}_k\}$ by minimizing an object function. Define $W(\mathcal{A},\mathcal{B}):= \sum_{v_i\in \mathcal{A}, v_j\in \mathcal{B}}w_{ij}$, $\bar{\mathcal{A}}$ the complement of $\mathcal{A}$, $vol(\mathcal{A}):= \sum_{v_i\in \mathcal{A}, v_j\in \mathcal{V}}w_{ij}$, and $|\mathcal{A}|$: the number of nodes in $\mathcal{A}$. There are two most commonly used object function in graph partitioning: 
\begin{equation}
    \begin{split}
        &\text{RatioCut}(\mathcal{A}_1, \dots, \mathcal{A}_k) := \frac{1}{2}\sum^k_{i=1}\frac{W(\mathcal{A}_i,\bar{\mathcal{A}}_i)}{|\mathcal{A}_i|}\\
        &\text{Ncut}:=\frac{1}{2}\sum^k_{i=1}\frac{W(\mathcal{A}_i,\bar{\mathcal{A}}_i)}{vol(\mathcal{A}_i)}
    \end{split}
\end{equation}
However, minimizing the RatioCut and Ncut object functions are NP-hard problems. Spectral clustering tries to solve an approximated problem of Them. The implication steps are shown in the following table:\par
\begin{table}[h]
\caption{Spectral Clustering Algorithm}
\hrule
\vspace{3pt}
\noindent Spectral Clustering
\vspace{3pt}
\hrule
\vspace{3pt}
\noindent\textbf{Input}: graph matrix $\mathbf{W}$ where $w_{ij}$ describes the connection intensity between node $v_i$ and node $v_j$.\\
\textbf{Output}: Bipartition $\mathcal{A}$ and $\bar{\mathcal{A}}$ of the input data. \\
1. Compute the diagonal degree matrix $\mathbf{D}$ with elements:
\begin{equation}
    d_i = \sum^n_{j=1}\mathbf{W}_{ij}
\end{equation}
2: Compute the normalized Laplacian matrix: 
\begin{equation}
    \mathbf{L}_{sym} = \mathbf{D}^{-\frac{1}{2}}(\mathbf{D}-\mathbf{W})\mathbf{D}^{-\frac{1}{2}}
\end{equation}
3:Compute the first $k$ eigenvectors $\{\mathbf{v}_1, \mathbf{v}_2, \dots, \mathbf{v}_k\}$ of $\mathbf{L}_{sym}$\\
4:Let $\mathbf{V}\in \mathbb{R}^{n\times k}$ be the matrix containing the vectors $\{\mathbf{v}_1, \mathbf{v}_2, \dots, \mathbf{v}_k\}$ as columns.\\
5:Form the matrix $\mathbf{T}\in\mathbb{R}^{n\times k}$ from $\mathbf{V}$ by normalizing the rows to norm 1. That is to set $t_{ij}=\frac{v_{ij}}{\sqrt{\sum_k v^2_{ik}}}$\\
6:For $i = 1\dots n$, let $\mathbf{y}_i\in \mathbb{R}^k$ be the vector corresponding to the $i$th row of $\mathbf{T}$.\\
7: Apply $k$ means algorithm on points $\{\mathbf{y}_1, \dotsm \mathbf{y}_i=n\}$ and get clusters $\{\mathcal{C}_1,\dots, \mathcal{C}_k\}$\\
8:Output clusters $\{\mathcal{A}_1, \dots, \mathcal{A}_k\}$ with $\mathcal{A}_i = \{j|\mathbf{y}_j\in\mathcal{C}_i\}$
\hrule
\vspace{3pt}
\end{table}

We only analyze the normalized Ncut with $k=2$ case. Other situations can be easily extended from this analysis.\par
For the object function $\min\limits_{\mathcal{A}\subset \mathcal{V}} Ncut(A,\bar{A})$, we define the vector $f=(f_1, \dots, f_n)^T \in \mathbb{R}^n$ with elements
\begin{equation}\label{f_form}
    f_i = \begin{cases}
    & \sqrt{\frac{vol(\bar{\mathcal{A}})}{{vol(\mathcal{A})}}} \quad \text{if } v_i\in \mathcal{A}\\
    & - \sqrt{\frac{vol(\bar{\mathcal{A}})}{{vol(\mathcal{A})}}} \quad \text{if } v_i\in \bar{\mathcal{A}}\\
    \end{cases}
\end{equation}
We can prove that $(\mathbf{D}f)^T\mathbbm{1} = 0, f^TDf = vol(\mathcal{V})$ and $f^T\mathbf{L}f = vol(\mathcal{V})Ncut(A,\bar{A})$. Thus we can rewrite the problem of minimizing Ncut by the following equation.
\begin{equation}
\min\limits_{A}f^T\mathbf{L}f \quad \text{s.t. } f \text{ in form eq.(\ref{f_form})}, \mathbf{D}f\perp \mathbbm{1}, f'^T\mathbf{D}f = vol(\mathcal{V})
\end{equation}
Then we relax this problem by allowing $f$ to take arbitrary real values. 
\begin{equation}
    \min_{f\in R^n} f^T\mathbf{L}f \text{s.t. } \mathbf{D}f\perp \mathbbm{1}, f^T\mathbf{D}f = vol(\mathcal{V})
\end{equation}
Then we substitute $g:= \mathbf{D}^{-\frac{1}{2}}\mathbf{LD}^{-\frac{1}{2}}$. After substitution, the problem is 
\begin{equation}\label{normalized_Ncut}
    \min\limits_{g\in R^n}g^T\mathbf{D}^{-\frac{1}{2}}\mathbf{LD}^{-\frac{1}{2}}g \text{ s.t. } g\perp \mathbf{D}^{\frac{1}{2}}\mathbbm{1}, ||g||^2 = vol(\mathcal{V})
\end{equation}
solution $g$ of eq.(\ref{normalized_Ncut}) is given by the second biggest eigenvector of $\mathbf{L}_{sym}$.

\section{Problem Formulation}\label{formula}
In multiple sample clustering, each clustering object is not a vector, but a set of vectors sampled from a hidden distribution. Suppose we have $n$ clustering objects $\{\mathcal{S}_1, \mathcal{S}_2,\dots, \mathcal{S}_n\}$ and each clustering object $\mathcal{S}_i$ is combined by $q_i$ multiple sample vectors $\mathcal{S}_i = \{\mathbf{a}_{i1}, \mathbf{a}_{i2}, \dots, \mathbf{a}_{iq_i}\}$. The vectors in set $\mathcal{S}_i$ are sampled from Gaussian distribution $\mu_i\sim \mathcal{N}(\mathbf{m}_i, \mathbf{\Sigma}_i)$. We can compute the estimated unbiased mean vector $\mathbf{\tilde{m}}_i$ and covariance matrix $\mathbf{\tilde{\Sigma}}_i$:
\begin{equation}
    \mathbf{\tilde{m}}_i = \frac{1}{q_i}\sum^{q_i}_{j=1}\mathbf{a}_{ij}
\end{equation}
\begin{equation}
    \mathbf{\tilde{\Sigma}}_i = \frac{1}{q_i-1}\sum^{q_i}_{j=1}(\mathbf{a}_{ij} - \mathbf{\tilde{m}}_i)(\mathbf{a}_{ij}-\mathbf{\tilde{m}}_i)^T
\end{equation}
For the traditional clustering algorithm, $k$-means, spectral clustering,etc, 
they can only use the mean vector $\{\mathbf{m}_1,\dots,\mathbf{m}_n\}$ to clustering vector groups. However, for Wasserstein distance based clustering, Bhattacharyya distance based clustering, KL divergence based clustering, and KL divergence++ algorithm, they can use the first moment information $\{\mathbf{m}_1,\dots,\mathbf{m}_n\}$, second moment information $\{\mathbf{\Sigma}_1, \dots, \mathbf{\Sigma}_n\}$, and even the whole distribution information into clustering. We will elaborate on these algorithms in section 4.

\section{Clustering Algorithm}
We are going to introduce a general framework for multiple samples clustering. Two algorithms: Wasserstein distance based clustering and Bhattacharyya distance based clustering is generated from the framework.\par
The structure of the framework is shown in the following figure.
\begin{figure}[h]
    \centering
    \includegraphics[width = \textwidth]{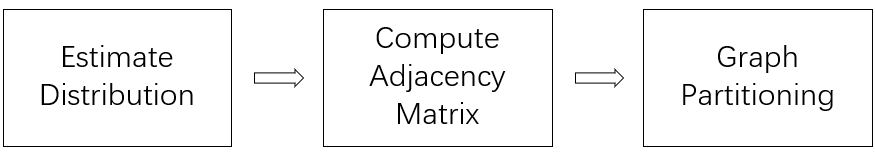}
    \caption{The Main Structure}
\end{figure}
We can see from the figure that there are three main parts of multiple sample clustering: estimating the distributions $\{\mu_1, \dots, \mu_n\}$ corresponded to each vector groups $\{\mathcal{S}_1, \dots, \mathcal{S}_n\}$, computing the adjacency matrix $\mathbf{W}$ based on a specific distribution metric, and using graph partitioning algorithms to output clustered groups $\{\mathcal{A}_1, \dots, \mathcal{A}_k\}$ where for $\forall \mathcal{S}_i\in\{\mathcal{S}_1, \dots, \mathcal{S}_n\}, \exists\mathcal{A}_j \in \{\mathcal{A}_1, \dots, \mathcal{A}_k\}$ satisfies that $\mathcal{S}_i\in\mathcal{A}_j \cup \mathcal{S}_i\notin \bar{\mathcal{A}}_j$. 
We summarize the general framework for multiple sample clustering in table \ref{framework}:
\begin{table}[h]
\caption{The general framework for multiple sample clustering}
\label{framework}
\hrule
\vspace{3pt}
\noindent Framework
\vspace{3pt}
\hrule
\vspace{3pt}
\noindent 1: $\{\mathcal{S}_1, \mathcal{S}_2,\dots, \mathcal{S}_n\}$ are the multiple sample groups\\
2: Estimate the hidden distribution $\mu_i$ from the $\mathcal{S}_i$\\
3: Compute the adjacency matrix $\mathbf{W}$ based on some distribution metric. $w_{ij} = D(\mu_i,\mu_j)$\\
4: Apply the graph partitioning algorithm on adjacency matrix $\mathbf{W}$ and output the clustering results $\{\mathcal{A}_1, \dots, \mathcal{A}_k\}$\\
\hrule
\end{table}
\subsection{Wasserstein Distance Based Clustering}
 The Wasserstein distance based clustering algorithm is a special case of the framework. For multiple samples dataset $\{\mathcal{S}_1, \dots, \mathcal{S}_n\}$ where $\mathcal{S}_i = \{a_{i1},\dots, a_{iq_i}\}$. We assume that each group of samples is drawn from the identical Gaussian distribution $\mu\sim \mathcal{N}(\mathbf{m}_i, \mathbf{\Sigma}_i)$ so that the mean vector $\{\mathbf{m}_1, \dots, \mathbf{m}_n\}$and covariance matrices $\{\mathbf{\Sigma}_1, \dots, \mathbf{\Sigma}_n\}$ are the  sufficient statistic of the Gaussian distribution $\{\mu_1, \dots, \mu_n\}$. Then we utilize the Wasserstein distance to build a symmetric adjacency matrix $\mathbf{W}: w_{ij} = D_{W,2}(\mu_i,\mu_j)$. Finally, we apply the normalized spectral clustering on the adjacency matrix to get the finally cluster $\{\mathcal{A}_1, \dots, \mathcal{A}_k\}$. The pseudo code of this algorithm is shown in table \ref{Wasserstein}.
 \begin{table}[h]
\caption{The Wasserstein Distance Based Clustering Algorithm}
\label{Wasserstein}
 \vspace{3pt}
\noindent\hrule
\vspace{3pt}
\noindent\textbf{Algorithm1}: The Wasserstein Distance Based Clustering
\vspace{3pt}
\hrule
\vspace{5pt}
\noindent
1: Gaussian distributions $\{\mu_1,\mu_2,\dots, \mu_n\}$.\\
2: $\{ \mathbf{m_1}, \mathbf{m_2}, \dots, \mathbf{m_n}\} \leftarrow$ means of input Gaussian distributions \\
3: $\{\mathbf{\Sigma_1}, \mathbf{\Sigma_2}, \dots, \mathbf{\Sigma_n}\} \leftarrow$ covariance matrices of input Gaussian distributions\\
\# Build the Wasserstein distance based graph matrix for the following spectral clustering\\
4: \textbf{for} i = 1:n \\
5: $\quad$ \textbf{for} j = 1:n\\
6: $\quad\quad \mathbf{X}_{ij} = D_{W,2}(\mu_i, \mu_j)$\\
7: $\quad$\textbf{end}\\
8: \textbf{end}\\
9: \# Do the normalization, transform the distance matrix $\mathbf{X}$ into the adjacency matrix $\mathbf{W}$, $max(\mathbf{X})$ output the biggest entry in $\mathbf{X}$\\
10: $\mathbf{W} = exp(\frac{-\mathbf{X}^2}{2\sigma^2})$ \quad $\mathbf{X}.^2$ compute the square of each entry in $\mathbf{X}$.\\
11: \# Apply the spectral clustering on the normalized matrix $\mathbf{W}$\\
12: $\{\mathcal{A}_1, \dots, \mathcal{A}_k\}$ = spectral clustering($\mathbf{W}$)
\hrule
\end{table}

\subsection{Bhattacharyya Distance Based Clustering}
 The Bhattacharyya distance based clustering algorithm has the same form of Wasserstein distance based clustering.  The only difference is that it replace the adjacency matrix $\mathbf{W}$ with $\mathbf{B}$ where $b_{ij} = D_B(\mu_i,\mu_j)$. The pseudo code of this algorithm is shown in table \ref{Bhattacharyya}.
 \begin{table}[h]
 \caption{Bhattacharyya Distance Based Clustering}
 \label{Bhattacharyya}
 \vspace{3pt}
\noindent \textbf{Algorithm 2:} Bhattacharyya Distance Based Clustering
\hrule
\vspace{3pt}
\hrule
\vspace{5pt}
\noindent
1: Gaussian distributions $\{\mu_1,\mu_2,\dots, \mu_n\}$.\\
2: $\{\mathbf{m_1}, \mathbf{m_2}, \dots, \mathbf{m_n}\} \leftarrow$ means of input Gaussian distributions \\
3: $\{\mathbf{\Sigma_1}, \mathbf{\Sigma_2}, \dots, \mathbf{\Sigma_n}\} \leftarrow$ covariance matrices of input Gaussian distributions\\
\# Build the Bhattacharyya distance matrix for the following spectral clustering\\
4: \textbf{for} i = 1:n \\
5: $\quad$ \textbf{for} j = 1:n\\
6: $\quad\quad \mathbf{X}(i,j) = D_B(\mu_i, \mu_j)$\\
7: $\quad$\textbf{end}\\
8: \textbf{end}\\
9: \# Do the normalization, transform the distance matrix $\mathbf{X}$ into the adjacency matrix $\mathbf{B}$, $max(\mathbf{X})$ output the biggest entry in $\mathbf{X}$\\
10: $\mathbf{B} = exp(\frac{-\mathbf{X}^2}{2\sigma^2})$ \quad $\mathbf{X}.^2$ compute the square of each entry in $\mathbf{X}$ themselves.\\
11: \# Apply the spectral clustering on the normalized matrix $\mathbf{B}$\\
12: $\{\mathcal{A}_1,\dots, \mathcal{A}_k\}$  spectral clustering($\mathbf{B}$)
\hrule
\end{table}

\subsection{KL-divergence based clustering Algorithm}
KL-divergence algorithm is purposed by Dhillon in 2007\cite{davis2007differential}. Unlike the previous algorithms only allowing scalar samples clustering, this algorithm firstly makes vector samples clustering possible. We can view this algorithm as an extension of $k$-means algorithm. The algorithm can be separated into three steps. Step one, for $n$ vector groups $\{\mathcal{S}_1, \dots, \mathcal{S}_n\}$, estimate the corresponded distribution $\{\mu_1,\dots, \mu_n\}$ where $\mu_i\sim\mathcal{N}(\mathbf{m}_i, \mathbf{\Sigma}_i)$, make the initial assignment $\pi^{(0)}_i = j \leftrightarrow \mathcal{S}_i \in \mathcal{A}_j$. Step two, in $t$-th loop, compute $k$ clustering centers $\{\nu_1^{(t)}, \dots, \nu_k^{(t)}\}$ $\nu_j^{(t)} = \underset{\nu\sim\mathcal{N}(\mathbf{m},\mathbf{\Sigma})}{\arg\min}\sum_{\{i:\pi^{(t)}_i = j\}} D_{KL}(\mu_{i}||\nu), j =1,\dots, k$. Step three, make an assignment $\pi^{(t)}$.  Vector group $\mathcal{S}_i$ is assigned to $\mathcal{A}_j$ if the corresponded distribution $\mu_i$ has the smallest KL-divergence with cluster center $\nu_j^{(t+1)} \leftrightarrow \pi^{(t)}_i = j$ if $\nu_j^{(t)} = \underset{\nu\in\{\nu_1^{(t)}, \dots, \nu_k^{(t)}\}}{\arg\min}D_{KL}(\mu_i||\nu)$. Come back to step 2 until the cluster assignment does not change. The pseudo code of this algorithm is shown in table \ref{KL divergence}.
\begin{table}[h]
\caption{KL divergence based clustering algorithm}
\label{KL divergence}
\hrule
\textbf{Algorithm 3:} KL divergence based clustering
\hrule
1: Input Gaussians $\{\mu_1,\mu_2,\dots, \mu_n\}$\\
2: $\{\mathbf{m}_1, \mathbf{m}_2, \dots, \mathbf{m}_n\}$ are the mean vectors of input Gaussians\\
3: $\{\mathbf{\Sigma}_1, \mathbf{\Sigma}_2,\dots, \mathbf{\Sigma}_n\}$ are the covariance matrices of input Gaussians\\
4: $\pi^{(0)} \leftarrow$ initial cluster assignments\\
5: \textbf{while} not converge \textbf{do}\\
6: \quad \textbf{for} $j=1:k$ \textbf{do}  \# update means of clustering centers $\{\nu_1, \dots, \nu_k\}, \quad \nu_i\sim\mathcal{N}(\mathbf{a}_i, \mathbf{X}_i)$\\
7: \quad \quad $\mathbf{a}_j\leftarrow \frac{1}{|\{i:\pi_i^{(t)}=j\}|}\sum_{i:\pi_i^{(t)}=j}\textbf{m}_i$\\
8: \quad \textbf{end for}\\
9: \quad \textbf{for} $j=1:k$ \textbf{do} \# update cluster covariances of clustering centers $\{\nu_1, \dots, \nu_k\}$\\
10: \quad \quad $\mathbf{X}_j\leftarrow\frac{1}{|\{i:\pi_i^{(t)}=j\}|}\sum_{i:\pi_i^{(t)}=j}(\mathbf{\Sigma}_i+(\mathbf{m}_i-\mathbf{a}_j)(\mathbf{m}_i - \mathbf{a}_j)^T)$\\
11: \quad \textbf{end for}\\
12: \quad \textbf{for} $i=1:n$ \textbf{do} \# assign each Gaussian to the closest cluster representative Gaussian\\
13: \quad \quad $\pi_i^{(t+1)}\leftarrow\underset{1\leq j\leq k }{\arg\min}D_{KL}(\mu_i||\nu_j)$\\
14: \quad \textbf{end for}\\
15: \textbf{end while}
\vspace{3pt}
\hrule
\end{table}

\subsection{KL divergence++ algorithm}
KL divergence++ is an improved version of KL-divergence based clustering algorithm. This algorithm is inspired by Arthur\cite{arthur2007k}. Arthur choose the initial clustering center carefully by making the distances between clustering centers statistically big enough\cite{arthur2007k}. Which yields KL divergence++ algorithm. 
Section 4.3 refers KL divergence based clustering algorithm has three steps. KL divergence++ only changes the step one. In step one, we do not make assignment $\pi$ randomly. Instead, we choose $k$ initial clustering centers in a sequence. The step one in KL divergence based clustering algorithm is separated into three steps. Step one, randomly choose the first clustering center $\nu_1^{(0)}$ from $\{\mu_1, \dots, \mu_n\}$. Step two, for the $i$-th$(i>1)$ clustering center, we need to compute the KL divergence between each distribution $\mu_{i}$ and $(i-1)$ chosen clustering center $\{D_{KL}(\mu_i,\mu_(1){r_1}, \dots, D_{KL}(\mu_i,\mu_(1){r_{m-1}})\}$. Then choose the smallest element of this vector $D_{KL,i}*$.\par. 
Compute the the probability vector $\{p_{1,m}, p_{2,m}, dots, p_{n,m}\}$. The $p_{i,m}$ is proportional to $D_{KL,i}^*$. Step Three,  Choose the $m$th clustering center based on the probability vector. go back to step two until all clustering center is chosen.
\hrule
\vspace{3pt}
\noindent
\textbf{Algorithm 4:} KL-divergence++ algorithm
\vspace{3pt}
\hrule
\vspace{3pt}
\noindent
\# Use k means++ algorithm to assign the initial cluster center\\
1: For vector groups $\{\mathcal{S}_1, \dots, \mathcal{S}_n\}$, estimate the corresponded Gaussian distributions:$\{\mu_1,\mu_2,\dots, \mu_n\}$\\
2: $\{\mathbf{m}_1, \mathbf{m}_2, \dots, \mathbf{m}_n\}$ are the mean vectors of input Gaussians\\
3: $\{\mathbf{\Sigma}_1, \mathbf{\Sigma}_2,\dots, \mathbf{\Sigma}_n\}$ are the covariance matrices of input Gaussians.\\
4: Choose the first clustering center $\nu^{(0)}_1$ randomly from $n$ input Gaussians.\\
5: \textbf{for} q = 1:(k-1) \# choose the $q+1$-th clustering center $\nu^{(0)}_{q+1}$\\ 
6: \quad Compute $n\times q$ divergence matrix $\mathbf{D}$ where $d_{ij} = D_{KL}(\mu_i||\nu^{(0)}_j)\quad i = 1,2,\dots, n, j = 1,2,\dots, q$\\
7: \quad Build the minimum distance vector $\{d_1^*, d_2^*,\dots, d_n^*\}$ where $d_i^* = min(\mathbf{D}_{i.})$.\\
8: \quad Compute the probability vector $\{p_{(1,q)}, p_{(2,q)}, \dots, p_{(n,q)}\}$ where $p_{(i,q)} = \frac{d_i^*}{\sum_{j=1}^n d_j^*}$\\
9: \quad Choose $\nu_{q+1}$in the range of $\{\mu_1, \mu_2, \dots, \mu_n\}$ based on probability vector $\{p_{(1,q)}, p_{(2,q)}, \dots, p_{(n,q)}\}$. \\
10: \textbf{end for}\\
11: Arrange the initial assignment $\pi$ based on the initial $k$ cluster center $\{\nu^{(0)}_1,\nu^{(0)}_2,\dots, \nu^{(0)}_k\}$.\\
\# Do the KL divergence based cluster\\
12: \textbf{while} not converge \textbf{do}\\
13: \quad \textbf{for} $j=1:k$ \textbf{do}  \# update means of clustering centers $\{\nu_1, \dots, \nu_k\}, \quad \nu_i\sim\mathcal{N}(\mathbf{a}_i, \mathbf{X}_i)$\\
14: \quad \quad $\mathbf{a}_j\leftarrow \frac{1}{|\{i:\pi_i^{(t)}=j\}|}\sum_{i:\pi_i^{(t)}=j}\textbf{m}_i$\\
15: \quad \textbf{end for}\\
16: \quad \textbf{for} $j=1:k$ \textbf{do} \# update cluster covariances of clustering centers $\{\nu_1, \dots, \nu_k\}$\\
17: \quad \quad $\mathbf{X}_j\leftarrow\frac{1}{|\{i:\pi_i^{(t)}=j\}|}\sum_{i:\pi_i^{(t)}=j}(\mathbf{\Sigma}_i+(\mathbf{m}_i-\mathbf{a}_j)(\mathbf{m}_i - \mathbf{a}_j)^T)$\\
18: \quad \textbf{end for}\\
19: \quad \textbf{for} $i=1:n$ \textbf{do} \# assign each Gaussian to the closest cluster representative Gaussian\\
20: \quad \quad $\pi_i^{(t+1)}\leftarrow\underset{1\leq j\leq k }{\arg\min}D_{KL}(\mu_i||\nu_j)$\\
21: \quad \textbf{end for}\\
22: \textbf{end while}
\vspace{3pt}
\hrule

\section{Simulation Results}
In this section, we will apply $k$ means, spectral clustering, Wasserstein distance based clustering, Bhattacharyya distance based clustering, KL divergence based clustering and KL divergence++ into synthetic data and stock price data. Compare their clustering accuracy on these data sets. We will directly compute the normalized mutual information, a clustering accuracy index, between the clustered result and ground truth in synthetic data. However, the accuracy of unsupervised learning is very hard to measure in real data set since we do not have the ground truth. As a result, in the stock price data set, we decide to use the clustering result of the original data set as the ground truth. Then, we will add different levels of i.i.d Gaussian noise on the data. Finally, compute the clustering accuracy by computing the mutual information between the clustering result of noised data set and "ground truth". We will firstly introduce mutual information in section 5.1 about the accuracy index in unsupervised learning: \textit{Mutual information}. Then elaborate the simulation process and result of synthetic data and the real stock price data.

\subsection{Normalized Mutual information}
Normalized mutual information(NMI) is an index estimating the clustering quality. It comes from the mutual information. Mutual information is a measure of the mutual dependence between two variables. More specifically, it quantifies the "amount of information" obtained about one random variable through observing the other random variable. Suppose we have two different clustering results $\pi_a$ and $\pi_b$ of $n$ clustering objects $\{\mathcal{S}_1, \dots, \mathcal{S}_n\}$. It is easy to transform $\pi_a$ and $\pi_b$ into two discrete variables $\mathbf{X}$ and $\mathbf{Y}$. Let $p_{\mathbf{X}}(x)$ and $p_{\mathbf{Y}}(y)$ be the corresponded distribution respectively defined on space $\mathcal{X}$ and $\mathcal{Y}$. Then we have 
\begin{equation}
    \begin{split}
        &p_{\mathbf{X}}(i) = \frac{1}{|\mathcal{X}|}|\pi_a = i|\\
        &p_{\mathbf{Y}}(j) = \frac{1}{|\mathcal{Y}|}|\pi_b = j|\\
        &p_{\mathbf{X,Y}}(i,j) = \frac{1}{|\mathcal{X}\times\mathcal{Y}|}|(\pi_a = i, \pi_b = j)|
    \end{split}
\end{equation}
$|\pi = j|$ is the number of clustering objects assigned to cluster $j$.
The mutual information between $\mathbf{X}$ and $\mathbf{Y}$ is defined as 
\begin{equation}
   I(\mathbf{X},\mathbf{Y}) = \sum\limits_{y\in\mathcal{Y}}\sum\limits_{x\in\mathcal{X}}p_{(\mathbf{X,Y})}(x,y)\log\frac{p_{(\mathbf{X,Y})}(x,y)}{p_{\mathbf{X}}(x)p_{\mathbf{Y}}(y)}
\end{equation}
Define the entropy of random variable $\mathbf{X}$ and $\mathbf{Y}$ as followed. Then we have 
\begin{equation}
\begin{split}
    &H(\mathbf{X}) = -\sum\limits_{x\in\mathcal{X}}p_{\mathbf{X}}(x)\log p_{\mathbf{X}}(x)\\
    &H(\mathbf{Y}) = -\sum\limits_{y\in\mathcal{Y}}p_{\mathbf{Y}}(y)\log p_{\mathbf{Y}}(y)
\end{split}
\end{equation}

The normalized mutual information is defined as 
\begin{equation}
    NMI(\mathbf{X},\mathbf{Y}) = \frac{2\times I(\mathbf{X},\mathbf{Y})}{H(\mathbf{X}) + H(\mathbf{Y})}
\end{equation}
\subsection{synthetic data}
\begin{figure}[h]
\begin{subfigure}{0.5\textwidth}
\centering
    \includegraphics[width = \linewidth]{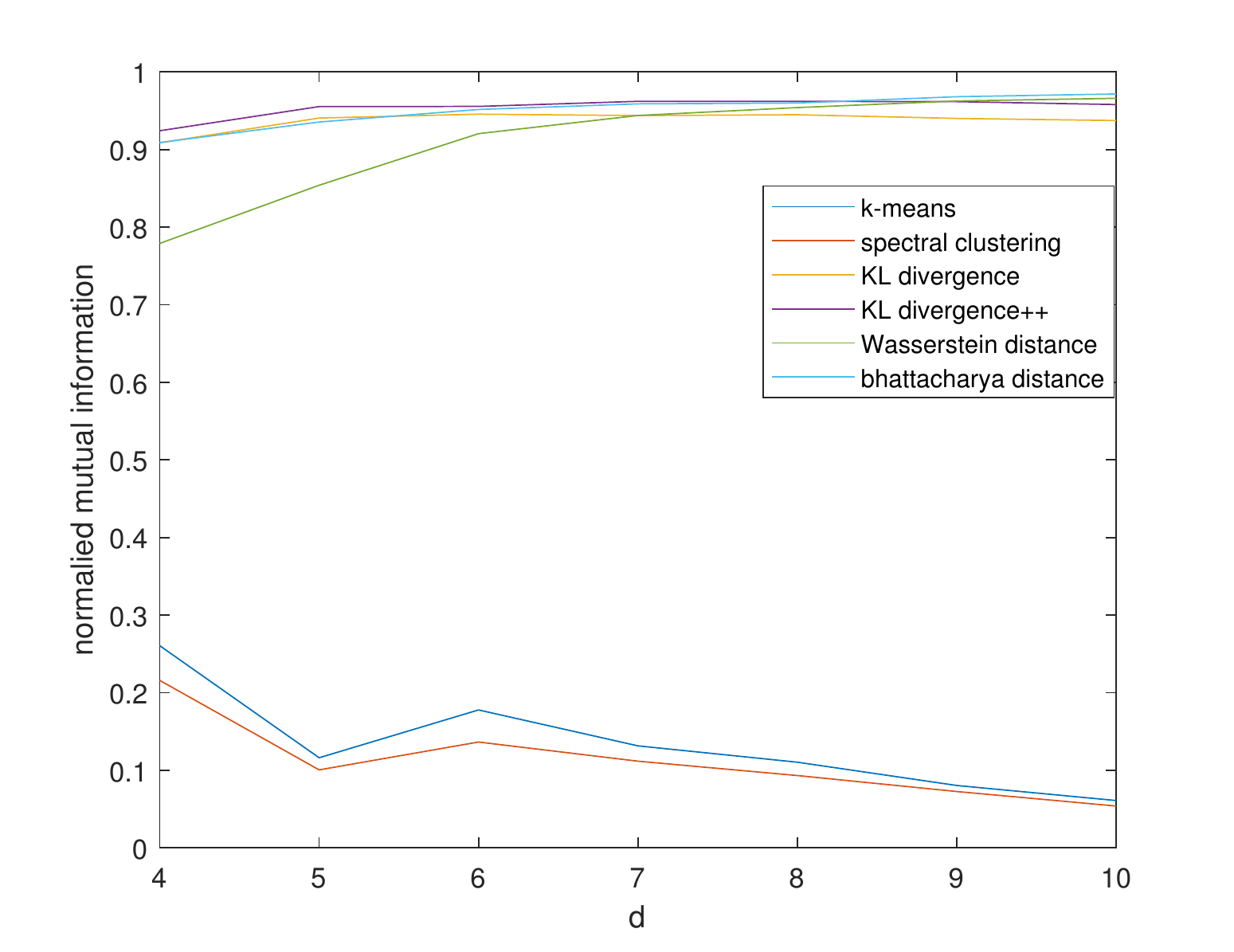}
    \caption{d = 4:10,k=5}
\end{subfigure}
\begin{subfigure}{0.5\textwidth}
\centering
    \includegraphics[width= \linewidth]{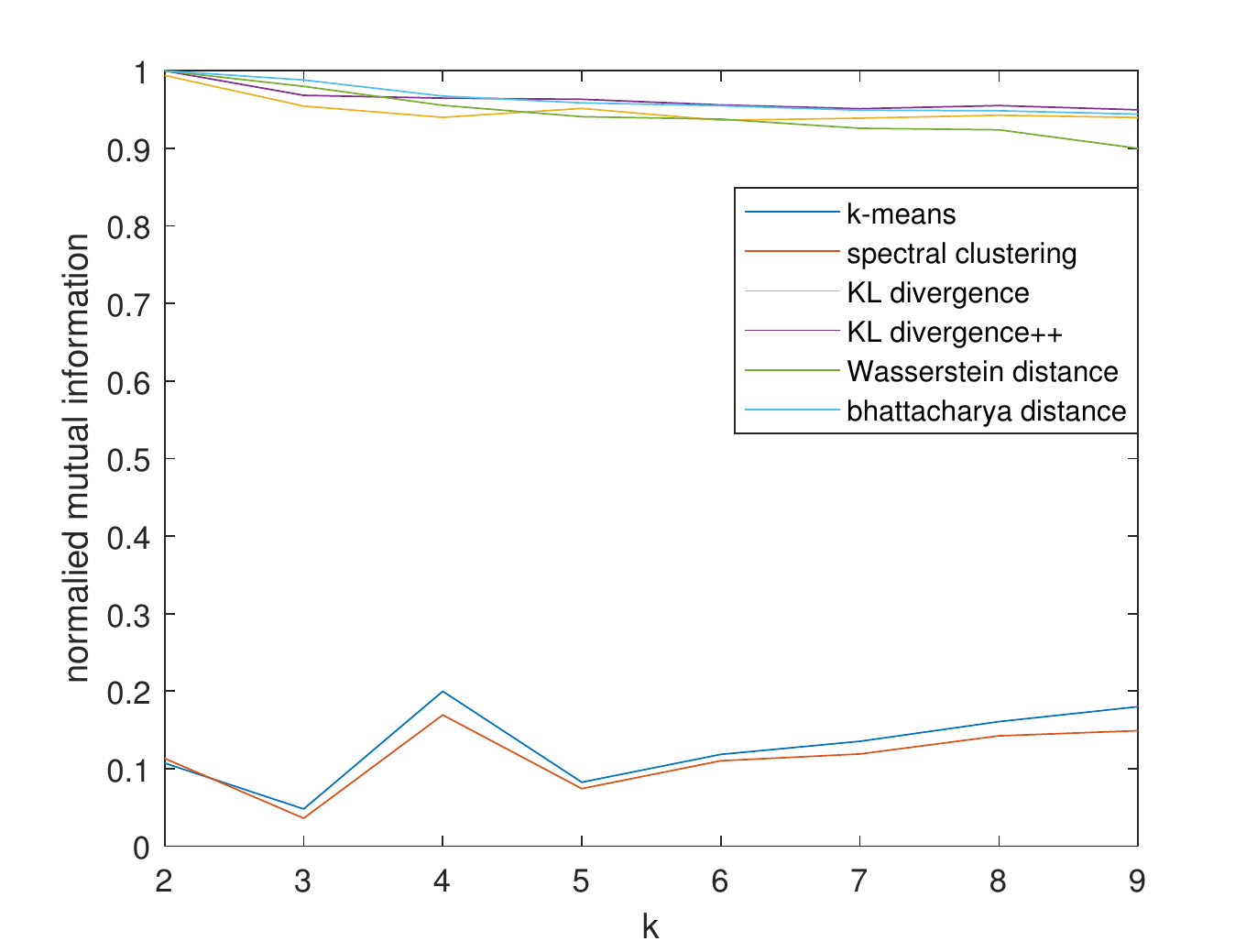}
    \caption{k = 2:9, d = 7}
\end{subfigure}
\caption{Synthetic Data}
\label{ave_syn}
\end{figure}

We apply the same synthetic data generating strategy as \cite{davis2007differential} to maintain comparability. Our synthetic dataset is a set of 200 objects. Each object consists 30 samples. The samples of the same object are drawn from same one of $k$ randomly generated $d$-dimensional multivariate Gaussians $\{\mu_1, \dots, \mu_k\}$. For $\mu_i(\mathbf{m}_i, \mathbf{\Sigma}_i)\in\{\mu_1, \dots, \mu_k\}$. The mean vector $\mathbf{m}_i$ is chosen uniformly at the surface of the $d$-dimensional unit simplex. Covariance matrix $\mathbf{\Sigma}_i = \mathbf{U}\mathbf{S}\mathbf{U}^T$ is a random matrix with eigenvalues $\mathbf{S} = \{1,2, \dots, d\}$ and $\mathbf{U}$ a random orthogonal matrix. We apply six different algorithms, $k$-means, spectral clustering, KL divergence based clustering algorithm, KL divergence++, Wasserstein distance based clustering, and Bhattacharyya distance based clustering. The simulation results can been shown as in figure \ref{ave_syn}. The first figure shows the NMI line if we fix the number of clusters $k=5$ and change dimension from 4 to 10. The second show the NMI line if we fix the dimension $d = 4$ and change the number of clusters $k$. Every figure is based on the average results after 1000 iterations. This simulation is run on the computer with Intel I5-8400 and takes 51142 seconds. The average NMI is shown in figure.\ref{ave_syn}, table.\ref{Ave_k_5} and table.\ref{Ave_d_7}.

\begin{table}[H]
    \centering
    \caption{The Average of Normalized Mutual Information when $k = 5$}
    \begin{tabular}{|l|c|c|c|c|c|c|c|}
    \hline
        \diagbox{Algorithm}{$d$} & 4 & 5 & 6 & 7 & 8 & 9 & 10\\ \hline
         $k$ means & 0.2605 & 0.1161 & 0.1777 & 0.1314 & 0.1103 & 0.0804 & 0.0610\\ \hline
         spectral clustering & 0.2158 & 0.1005 & 0.1363 & 0.1117 & 0.0932 & 0.0726 & 0.0539 \\ \hline
         KL divergence & 0.9086 & 0.9406 & 0.9455 & 0.9438 & 0.9448 & 0.9401 & 0.9374\\ \hline
         KL divergence++ & 0.9242 & 0.9553 & 0.9555 & 0.9621 & 0.9621 & 0.9617 & 0.9580\\ \hline
         Wasserstein & 0.7790 & 0.8540 & 0.9204 & 0.9439 & 0.9540 & 0.9625 & 0.9660\\ \hline
         Bhattacharyya & 0.9090 & 0.9354 & 0.9516 & 0.9588 & 0.9599 & 0.9680 & 0.9716\\
    \hline
    \end{tabular}
    \label{Ave_k_5}
\end{table}

\begin{table}[H]
\centering
\caption{The Average of Normalized Mutual Information $d = 7$}
\begin{tabular}{|l|c|c|c|c|c|c|c|c|}
    \hline
    \diagbox{Algorithm}{$k$} & 2 & 3 & 4 & 5 & 6 & 7 & 8& 9\\ \hline
    $k$ means & 0.1073 & 0.0481 & 0.1998 & 0.0826 & 0.1185 & 0.1354 & 0.1607 & 0.1800\\ \hline
    spectral clustering & 0.1133 & 0.0361 & 0.1693 & 0.0743 & 0.1102 & 0.1190 & 0.1424 & 0.1490\\ \hline
    KL divergence & 0.9940 & 0.9545 & 0.9399 & 0.9515 & 0.9362 & 0.9389 & 0.9427 & 0.9397\\ \hline
    KL divergence++ & 1 & 0.9684 & 0.9647 & 0.9633 & 0.9560 & 0.9511 & 0.9551 & 0.9497\\ \hline 
    Wasserstein & 0.9999 & 0.9799 & 0.9555 & 0.9408 & 0.9377 & 0.9259 & 0.9239 & 0.8999\\ \hline
    Bhattacharyya & 1 & 0.9882 & 0.9674 & 0.9586 & 0.9549 & 0.9492 & 0.9485 & 0.9441\\ 
    \hline
    \end{tabular}
    \label{Ave_d_7}
\end{table}

\begin{figure}[H]
    \begin{subfigure}{.5\textwidth}
    \centering
    \includegraphics[width = \linewidth]{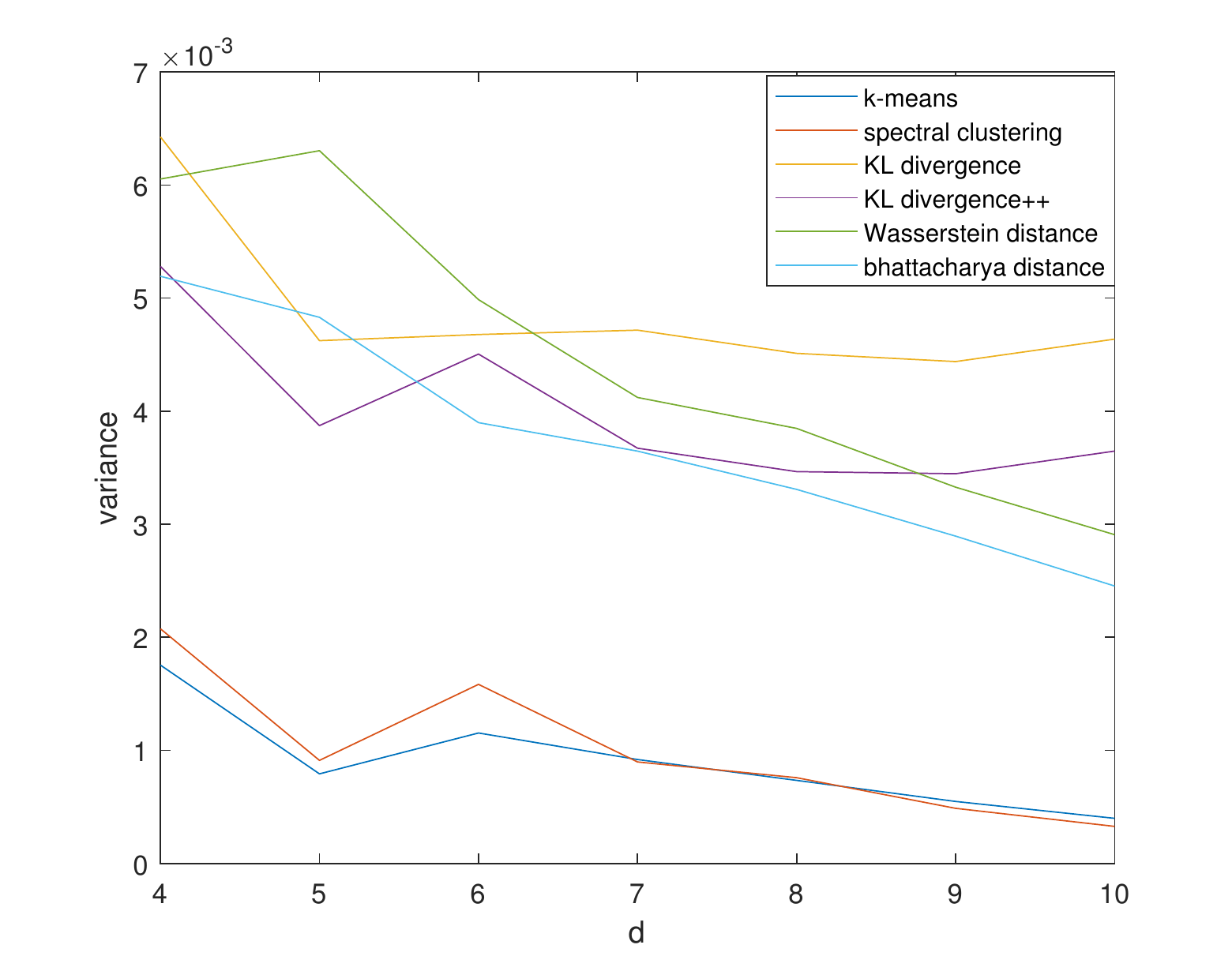}
    \caption{$d = 4:10, k = 5$}
    \end{subfigure}
    \begin{subfigure}{.5\textwidth}
    \centering
    \includegraphics[width = \linewidth]{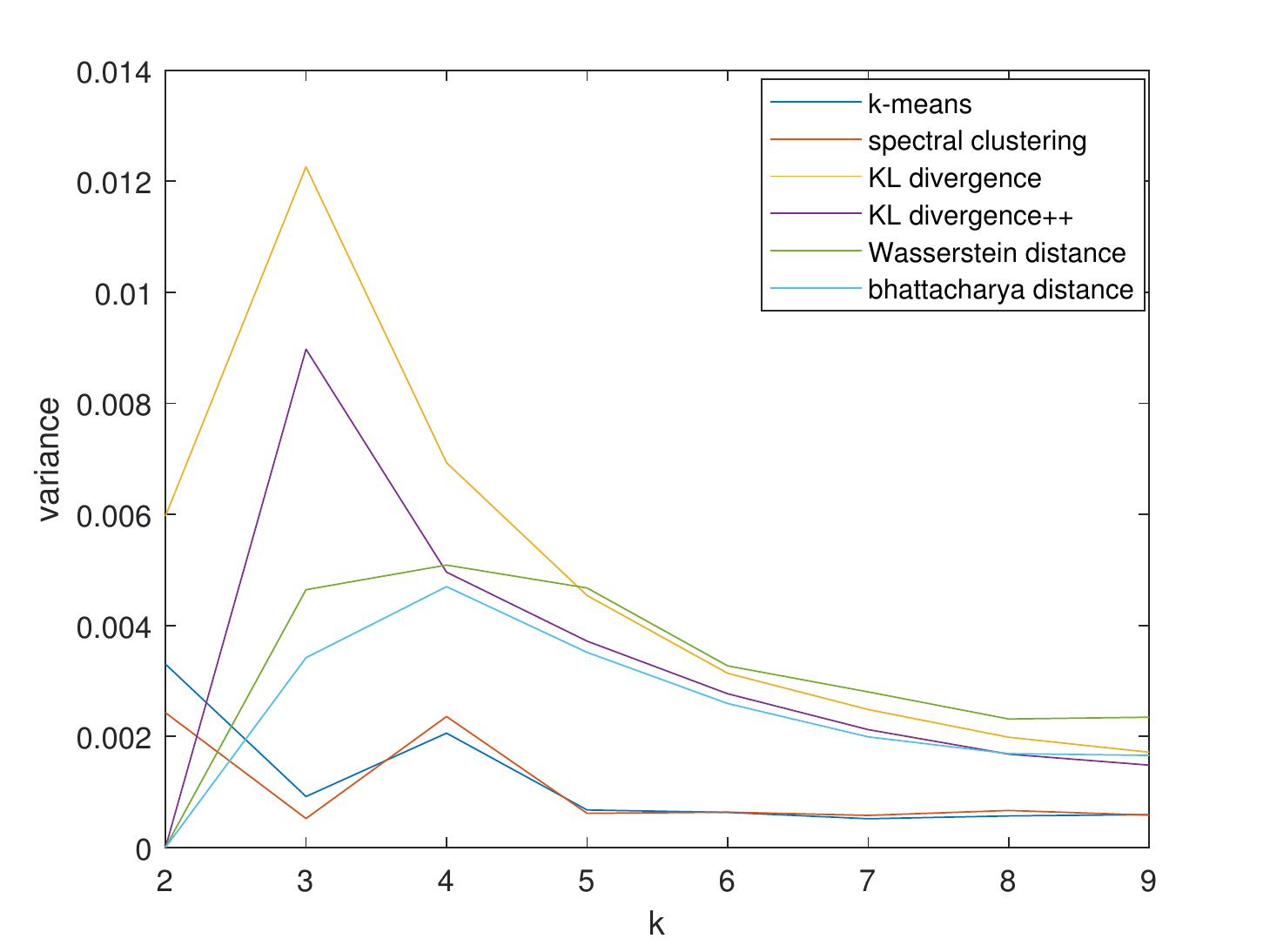}
    \caption{$k = 2:9, d = 7$}
    \end{subfigure}
    \caption{Variance of Normalized Mutual Information after 1000 iteration}
    \label{var_nmi}
\end{figure}

Then, we compute the variance of normalized mutual information for each algorithm after 1000 iteration. The detail data is shown in fig.\ref{var_nmi}, table.\ref{Var_k_5} and table.\ref{Var_d_7}.

\begin{table}[H]
    \centering
    \caption{The Covariance of Normalized Mutual Information in 1000 loops $k = 5$}
    \begin{tabular}{|l|c|c|c|c|c|c|c|}
    \hline
    \diagbox{Algorithm}{Dimension\\ $d$} & 4 & 5 & 6 & 7 & 8 & 9 & 10\\ \hline
        $k$ means & 0.0018 & 0.0008 & 0.0012 & 0.0009 & 0.0007 & 0.0005 & 0.0004\\ \hline
        spectral clustering & 0.0021 & 0.0009 & 0.0016 & 0.0009 & 0.0008 & 0.0005 & 0.0003\\ \hline
        KL divergence & 0.0064 & 0.0046 & 0.0047 & 0.0047 & 0.0045 & 0.0044 & 0.0046\\ \hline
        KL divergence++ & 0.0053 & 0.0039 & 0.0045 & 0.0037 & 0.0035 & 0.0034 & 0.0036\\ \hline
        Wasserstein & 0.0061 & 0.0063 & 0.0050 & 0.0041 & 0.0038 & 0.0033 & 0.0029\\ \hline
        Bhattacharyya & 0.0052 & 0.0048 & 0.0039 & 0.0036 & 0.0033 & 0.0029 & 0.0025\\
        \hline
    \end{tabular}
    \label{Var_k_5}
\end{table}

\begin{table}[H]
\caption{The Covariance of Normalized Mutual Information in 1000 loops $d = 7$}
\centering
\begin{tabular}{|l|c|c|c|c|c|c|c|c|}
    \hline
    \diagbox{Algorithm}{k} & 2 & 3 & 4 & 5 & 6 & 7 & 8 & 9\\ \hline
    $k$ means & 0.0033 & 0.0009 & 0.0021 & 0.0007 & 0.0007 & 0.0005 & 0.0006 & 0.0006\\ \hline
    spectral clustering & 0.0024 & 0.0005 & 0.0024 & 0.0006 & 0.0006 & 0.0006 & 0.0007 & 0.0006\\ \hline
    KL divergence & 0.0060 & 0.0123 & 0.0069 & 0.0045 & 0.0031 & 0.0025 & 0.0020 & 0.0017\\ \hline
    KL divergence++ & 0 & 0.0090 & 0.0050 & 0.0037 & 0.0028 & 0.0021 & 0.0017 & 0.0015\\ \hline
    Wasserstein & 0 & 0.0045 & 0.0051 & 0.0047 & 0.0033 & 0.0028 & 0.0023 & 0.0023\\ \hline
    Bhattacharyya & 0 & 0.0034 & 0.0047 & 0.0035 & 0.0026 & 0.0020 & 0.0017 & 0.0017\\
    \hline
\end{tabular}
\label{Var_d_7}
\end{table}

Compare the results between the first-moment information based clustering algorithms: $k$ means, spectral clustering and the distribution information based clustering algorithms: KL divergence based clustering, KL divergence++, Wasserstein distance based clustering, and Bhattacharyya distance based clustering. We can find the distribution information can greatly improve the clustering accuracy, and Bhattacharyya distance based algorithm has the highest average NMI. What's more, the Bhattacharyya distance based clustering also has the lowest variance among all second-moment information based algorithms. This simulation only use the Gaussian distribution as the assumed hidden distribution, but this results can be easily extended to more distribution assumptions. 

\subsection{Stock clustering}
We utilize the New York stock exchange data collected by Dominik Gawlik \cite{anudc:4896}. There are four different prices for each stock every day: open price(open), close price(close), low price (low) and high price (high). We can view each stock as a set of multiple samples on four features at different time. More specifically, 1726 $4$-dimension samples from date Jun 4th, 2010 to Oct 7th, 2016. Then do the clustering based on the samples. Six different algorithms are applied: $k$ means++, spectral clustering, KL divergence based clustering, KL divergence++, Wasserstein distance based spectral clustering and Bhattacharyya distance based clustering. Use the assignment result of these each algorithm as ground truth and then add i.i.d Gaussian noise $\mathcal{N}(\mu = 0,\sigma = 1.0),\mathcal{N}(\mu=0,\sigma=2.0),\mathcal{N}(\mu=0,\sigma=3.0)$ to the stock data. Apply the clustering algorithms to the noised data and compute the \textit{mutual information} with the 'ground truth' with respect to that algorithm. This simulation is run on platform I5-8400. 100 iteration takes 78450 seconds. We get the following six figures \ref{nyse}:\newpage
    \begin{figure}[H]
        \begin{subfigure}{0.5\textwidth}
            \centering
            \includegraphics[width = \linewidth]{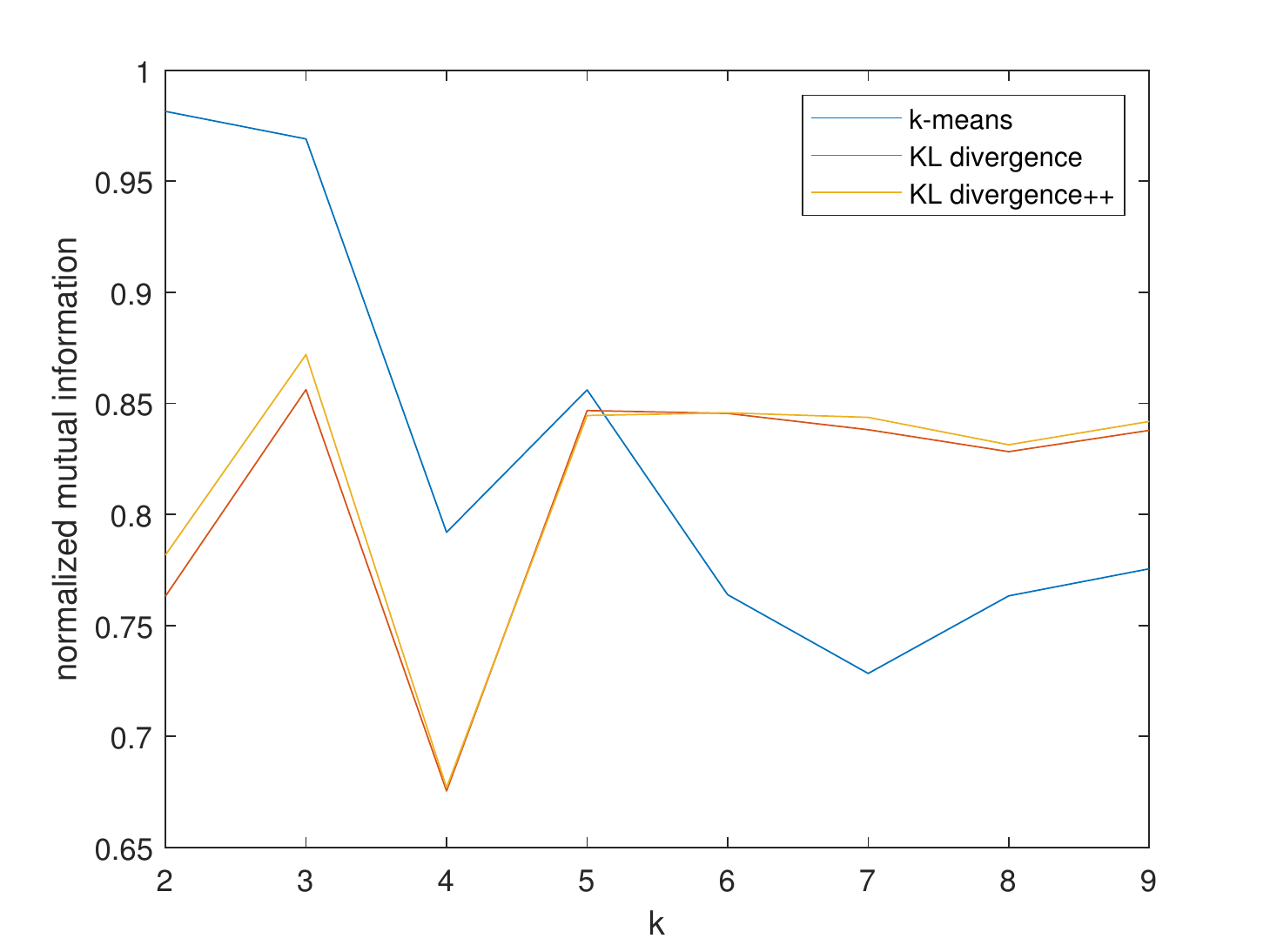}
            \caption{$k$ means based algorithm with $i.i.d$ Gaussian noise $\sigma = 1.0$}
        \end{subfigure}
        \begin{subfigure}{0.5\textwidth}
            \centering
            \includegraphics[width = \linewidth]{noise1_k_means.pdf}
            \caption{spectral clustering based algorithm with $i.i.d$ Gaussian noise $\sigma = 1.0$}
        \end{subfigure}
        
        \medskip
        \begin{subfigure}{0.5\textwidth}
            \centering
            \includegraphics[width = \linewidth]{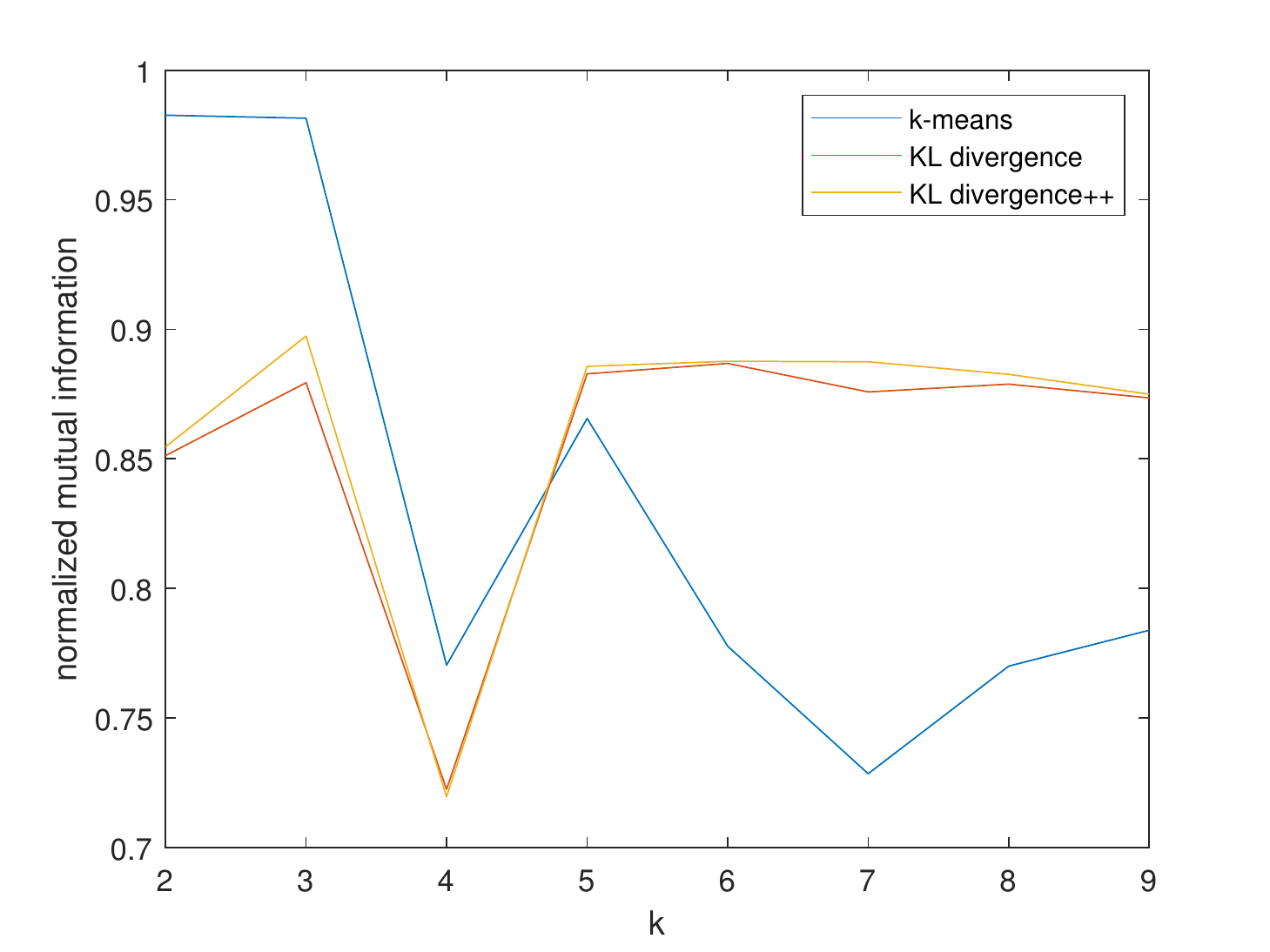}
            \caption{$k$ means based algorithm with $i.i.d$ Gaussian noise $\sigma = 2.0$}
        \end{subfigure}
        \begin{subfigure}{0.5\textwidth}
            \centering
            \includegraphics[width = \linewidth]{noise2_k_means.pdf}
            \caption{spectral clustering based algorithm with $i.i.d$ Gaussian noise $\sigma = 2.0$}
        \end{subfigure}
    
        \medskip
        \begin{subfigure}{0.5\textwidth}
            \centering
            \includegraphics[width = \linewidth]{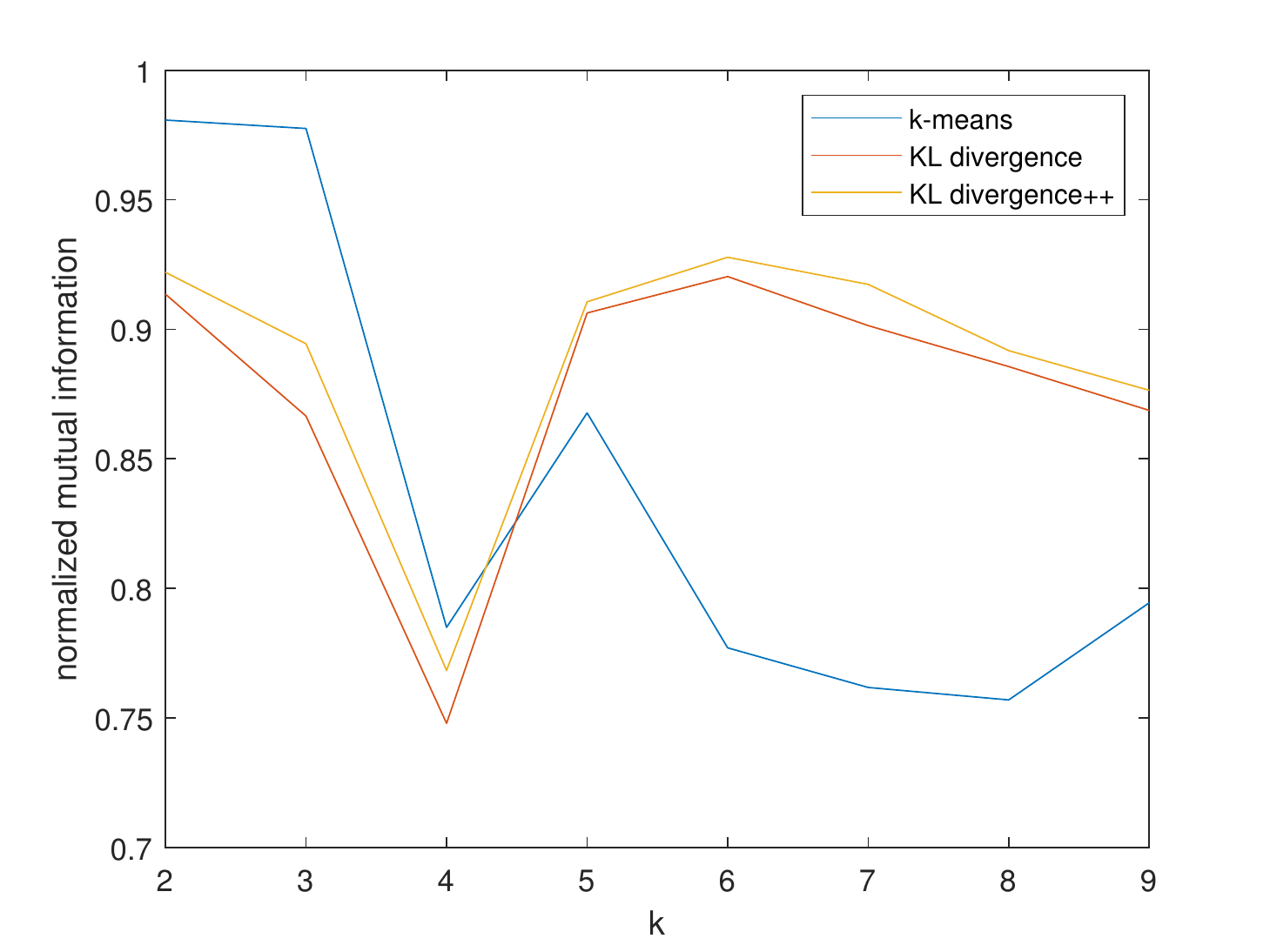}
            \caption{$k$ means based algorithm with $i.i.d$ Gaussian noise $\sigma = 3.0$}
        \end{subfigure}
        \begin{subfigure}{0.5\textwidth}
            \centering
            \includegraphics[width = \linewidth]{noise3_k_means.pdf}
            \caption{spectral clustering based algorithm with $i.i.d$ Gaussian noise $\sigma = 3.0$}
        \end{subfigure}
        \caption{New York Stock Data Set}
        \label{nyse}
    \end{figure}
    
    \begin{table}[H]
        \centering
        \caption{The average of Normalized Mutual Information $\sigma = 1.0$}
        \begin{tabular}{|l|c|c|c|c|c|c|c|c|}
        \hline
            \diagbox{Algorithm}{k} & 2 & 3 & 4 & 5 & 6 & 7 & 8 & 9\\ \hline
            $k$ means & 0.9815 & 0.9691 & 0.7919 & 0.8561 & 0.7639 & 0.7284 &  0.7633 & 0.7755\\ \hline
            spectral clustering & 0.9935 & 0.3093 & 0.3046 & 0.2835 & 0.1836 & 0.1335 & 0.1114 & 0.1076\\ \hline
            KL divergence & 0.7632 & 0.8562 & 0.6755 & 0.8467 & 0.8455 & 0.8381 & 0.8283 & 0.8378\\ \hline
            KL divergence++ & 0.7817 & 0.8720 & 0.6770 & 0.8445 & 0.8457 & 0.8437 & 0.8314 & 0.8419\\ \hline
            Wasserstein & 0.5979 & 0.7104 & 0.7659 & 0.6242 & 0.5761 & 0.1991 & 0.2655 & 0.3724\\ \hline
            Bhattacharyya & 0.6348 & 0.4759 & 0.7517 & 0.5741 & 0.5999 & 0.6229 & 0.6437 & 0.6489\\ \hline
        \end{tabular}
        \label{nyse_noise1}
    \end{table}
    
    \begin{table}[H]
        \centering
        \caption{The average of Normalized Mutual Information $\sigma = 2.0$}        
        \begin{tabular}{|l|c|c|c|c|c|c|c|c|}

        \hline
            \diagbox{Algorithm}{k} & 2 & 3 & 4 & 5 & 6 & 7 & 8 & 9\\ \hline
            $k$ means & 0.9826 & 0.9815 & 0.7704 & 0.8656 & 0.7777 & 0.7285 & 0.7700 & 0.7839\\ \hline
            spectral clustering 0.9935 & 0.3714 & 0.3305 & 0.2982 & 0.1891 & 0.1345 & 0.1066 & 0.0979\\ \hline
            KL divergence & 0.8513 & 0.8794 & 0.7225 & 0.8828 & 0.8868 & 0.8759 & 0.8789 & 0.8736\\ \hline
            KL divergence++ & 0.8547 & 0.8974 & 0.7196 & 0.8857 & 0.8877 & 0.8875 & 0.8826 & 0.8750\\ \hline
            Wasserstein & 0.5909 & 0.6411 & 0.7549 & 0.6587 & 0.6370 & 0.6628 & 0.6635 & 0.5387\\ \hline
            Bhattacharyya & 0.8677 & 0.6650 & 0.7683 & 0.6104 & 0.5979 & 0.6634 & 0.6426 & 0.6455\\ \hline
        \end{tabular}
        \label{nyse_noise2}
    \end{table}
    
    \begin{table}[H]
        \centering
        \caption{The average of Normalized Mutual Information $\sigma = 3.0$}
        \begin{tabular}{|l|c|c|c|c|c|c|c|c|}
        \hline
            \diagbox{Algorithm}{k} & 2 & 3 & 4 & 5 & 6 & 7 & 8 & 9\\ \hline
            $k$ means & 0.9808 & 0.9775 & 0.7850 & 0.8677 & 0.7771 & 0.7618 & 0.7570 & 0.7945\\ \hline
            spectral clustering & 0.9922 & 0.3920 & 0.3181 & 0.3040 & 0.1804 & 0.1270 & 0.0967 & 0.0951\\ \hline
            KL divergence & 0.9136 & 0.8665 & 0.7479 & 0.9063 & 0.9204 & 0.9014 & 0.8856 & 0.8687\\ \hline
            KL divergence++ & 0.9220 & 0.8944 & 0.7683 & 0.9106 & 0.9278 & 0.9174 & 0.8918 & 0.8765\\ \hline
            Wasserstein & 0.7037 & 0.6652 & 0.7387 & 0.7661 & 0.6339 & 0.7127 & 0.6890 & 0.6364\\ \hline
            Bhattacharyya & 0.8336 & 0.8843 & 0.7347 & 0.5583 & 0.6099 & 0.6337 & 0.6233 & 0.6457\\ \hline
        \end{tabular}
        \label{nyse_noise3}
    \end{table}
    
     In stock dataset simulation, we can split the algorithms into two groups: the $k$ means based algorithms: $k$ means++, KL divergence based clustering and KL divergence++ and spectral clustering based algorithms: spectral clustering, Wasserstein distance based spectral clustering, and Bhattacharyya distance based clustering. In each group, we can see that the introduction of second order information greatly improves the clustering stability. 

\section{Conclusion}
In this paper, we propose a framework for multiple sample clustering, this framework can solve all of the problems without data structure limitation. We propose two specific cases of this framework: Wasserstein distance based clustering and Bhattacharyya distance based clustering. We also proposed an improved KL divergence based clustering algorithm: KL divergence++. In synthetic data, the simulation  results show that introducing distribution information can greatly improves the clustering accuracy. In stock price data, distribution information can strengthen the stability of clustering results. This framework can be applied in different areas considering the pervasiveness of multiple samples data.
\printbibliography
\end{document}